\documentclass[a4paper,10pt]{article}
\usepackage{amsmath}
\usepackage[utf8]{inputenc}
\usepackage{float}
\usepackage{todonotes}
\usepackage{hyperref}
\usepackage{color,soul}
\usepackage{booktabs}
\usepackage{authblk}

\usepackage{graphicx}
\usepackage{subcaption}
\usepackage{tgbonum}

\newcommand\norm[1]{\left\lVert#1\right\rVert}

\title{Federated Learning in Distributed Medical Databases: Meta-Analysis of Large-Scale Subcortical Brain Data}
\author[1]{Santiago Silva}
\author[2]{Boris A. Gutman}
\author[3]{Eduardo Romero}
\author[4]{Paul M. Thompson}
\author[5]{Andre Altmann}
\author[1]{Marco Lorenzi}
\author[*]{for ADNI, PPMI, and UK Biobank}

\affil[1]{Université Côte d'Azur, Inria Sophia Antipolis, Epione Research Project, France}
\affil[2]{Department of Biomedical Engineering, Illinois Institute of Technology, USA}
\affil[3]{CIM@LAB, Universidad Nacional de Colombia, Bogotá, Colombia}
\affil[4]{USC Stevens Institute for Neuroimaging and Informatics, Los Angeles, USA}
\affil[5]{Centre for Medical Image Computing, UCL, London, UK}
\date{}

\begin{document}
\maketitle
\begin{abstract}
	At this moment, databanks worldwide contain brain images of previously unimaginable numbers. Combined with developments in data science, these massive data provide the potential to better understand the genetic underpinnings of brain diseases. However, different datasets, which are stored at different institutions, cannot always be shared directly due to privacy and legal concerns, thus limiting the full exploitation of big data in the study of brain disorders. Here we propose a \textit{federated learning} framework for securely accessing and meta-analyzing any biomedical data without sharing individual information. We illustrate our framework by investigating brain structural relationships across diseases and clinical cohorts. The framework is first tested on synthetic data and then applied to multi-centric, multi-database studies including ADNI, PPMI, MIRIAD and UK Biobank, showing the potential of the approach for further applications in distributed analysis of multi-centric cohorts. \\
	
	\textbf{Keywords:} Federated learning, distributed databases, PCA, SVD, meta-analysis, brain disease.
\end{abstract}

\section{Introduction}
\label{sec:intro}
Nowadays, a large amount of magnetic resonance images (MRI) scans are stored across a vast number of clinical centers and institutions. Researchers are currently analyzing these large datasets to understand the underpinnings of brain diseases. However, due to privacy concerns and legal complexities, data hosted in different centers cannot always be directly shared. In practice, data sharing is also hampered by the need to transfer large volumes of biomedical data with the associated bureaucratic burden. This situation led researchers to look for an analysis solution within  \textit{meta-analysis} or  \emph{federated learning} paradigms. In the federated setting, a model is fitted without sharing individual information across centres, but only model parameters.  Meta-analysis instead performs statistical testing by combining results from several independent assays\cite{Truffert2017, Sun2018}, for example by sharing \textit{p}-values, effect sizes, and/or standard errors across centers.

\let\thefootnote\relax\footnote{*Data used in preparation of this article were obtained from the Alzheimer’s Disease Neuroimaging Initiative (ADNI) database (adni.loni.usc.edu). As such, the investigators within the ADNI contributed to the design and implementation of ADNI and/or provided data but did not participate in analysis or writing of this report. A complete listing of ADNI investigators can be found at: \url{http://adni.loni.usc.edu/wp-content/uploads/how_to_apply/ADNI_ Acknowledgement_List.pdf}. 
Also, from the Parkinson’s Progression Markers Initiative (PPMI) database (www.ppmi-info.org/data). For up-to-date information on the study, visit \url{www.ppmi-info.org}. This research has been conducted using
the UK Biobank Resource.}

One of the best examples of such a research approach is the \textit{Enhancing NeuroImaging Genetics through Meta-Analysis} (ENIGMA) consortium (enigma.usc.edu). With a large number of institutions worldwide \cite{Thompson2014}, ENIGMA has become one of the largest networks bringing together multiple groups analyzing neuroimaging data from over 10,000 subjects. However, most of ENIGMA's secure meta-analytic studies in neuroimaging are performed using mass-univariate models.

The main drawback of mass-univariate analysis is that they can only model a single dependent variable at a time. This is a limiting assumption in most of the biomedical scenarios (e.g., neighboring voxels or genetic variations are highly correlated). To overcome this problem, multivariate analysis methods have been proposed to better account for covariance in high-dimensional data.

In a federated analysis context, a few works proposed generalization of standard neuroimaging multivariate analysis methods, such as Independent Component Analysis \cite{baker2015large}, sparse regression, and parametric statistical testing \cite{ming2017coinstac, plis2016}. Since these methods are mostly based on stochastic gradient descent, a large-number of communications across centers may be required to reach convergence. Therefore, there is a risk of computational and practical bottlenecks when applied to multi-centric high-dimensional data.

Lorenzi \textit{et} al.\cite{Lorenzi2017, Lorenzi2018} proposed a multivariate dimensionality reduction approach based on eigen-value decomposition. This approach does not require iteration over centers, and was demonstrated on the analysis of the joint variability in imaging-genetics data. However, this framework is still of limited practical utility in real applications, as data harmonization (e.g., standardization and covariate adjustment) should be also consistently performed in a federated way.

Herein we contribute to the state-of-the-art in federated analysis of neuroimaging data by proposing an end-to-end framework for data standardization, confounding factors correction, and multivariate analysis of variability of high-dimensional features. To avoid the potential bottlenecks of gradient-based optimization, the framework is based on schemes analysis through \textit{Alternating Direction Method of Multipliers} (ADMM) reducing the amount of iterations.

We illustrate the framework leveraging on the ENIMGA Shape tool, to provide a first application of federated analysis compatible with the standard ENIGMA pipelines.  It should be noted that, even though this work is here illustrated for the analysis of subcortical brain changes in neurological diseases, it can be extended to general multimodal multivariate analysis, such as to imaging-genetics studies.

The framework is benchmarked on synthetic data (section~\ref{sec:exp_synthetic}). It is then applied to the analysis of subcortical thickness and shape features across diseases from multi-centric, multi-database data including: Alzheimer's disease (AD), progressive and non-progressive mild cognitive impairment (MCIc, MCInc), Parkinson's disease (PD) and healthy individuals (HC) (section~\ref{sec:exp_real}).

\section{Methods}
\label{sec:methods}
Biomedical data is assumed to be partitioned across different centers restricting the access to individual information. However, centers can individually share model parameters and run pipelines for feature extraction. 


We denote the \emph{global} data (e.g., image arrays) and covariates (e.g., age, sex information) as respectively $\mathbf{X}$ and $\mathbf{Y}$,  obtained by concatenating respectively data and covariates of each center. Although these data matrices cannot be computed in practice, this notation will be used to illustrate the proposed methodology. In the global setting, variability analysis can be performed by analyzing the \textit{global data covariance matrix} $\mathbf{S}$.

For each center $c \in \{ 1,\dots, C \}$ with $N_c$ subjects each, we denote by $\mathbf{X}_c=(\mathbf{x}_i )_{i=1}^{N_c}$ and $\mathbf{Y}_c=(\mathbf{y}_i)_{i=1}^{N_c}$ the \emph{local} data and covariates. The feature-wise mean and standard deviation vectors of each center are denoted as $\Bar{\mathbf{x}}_c$ and $\mathbf{\sigma}_c$.

\begin{figure}
    \centering
    \begin{subfigure}[b]{0.32\linewidth}
        \includegraphics[width=\linewidth]{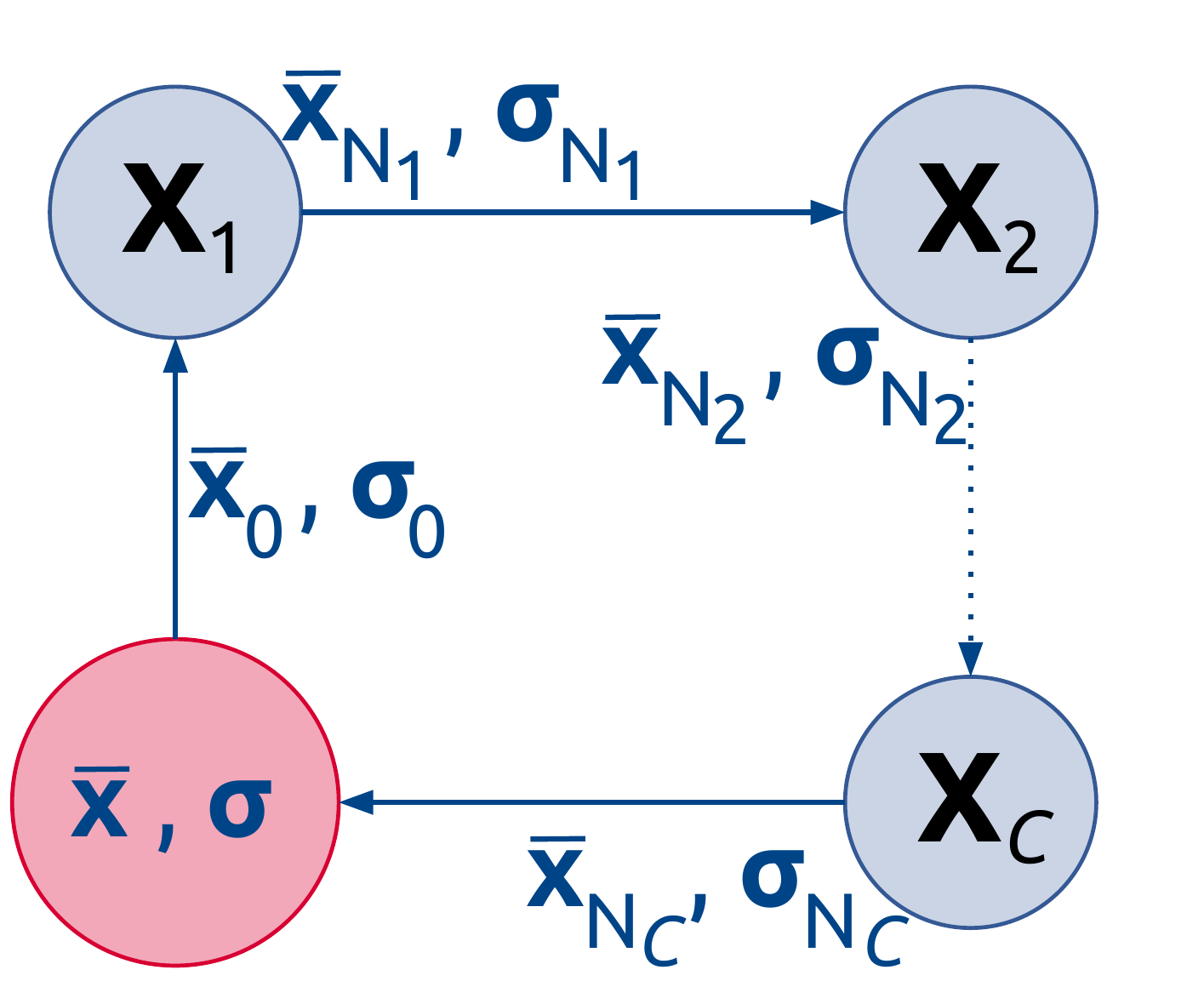}
        \caption{}
        \label{fig:pipeline_welford}
    \end{subfigure}
    \begin{subfigure}[b]{0.32\linewidth}
        \includegraphics[width=\linewidth]{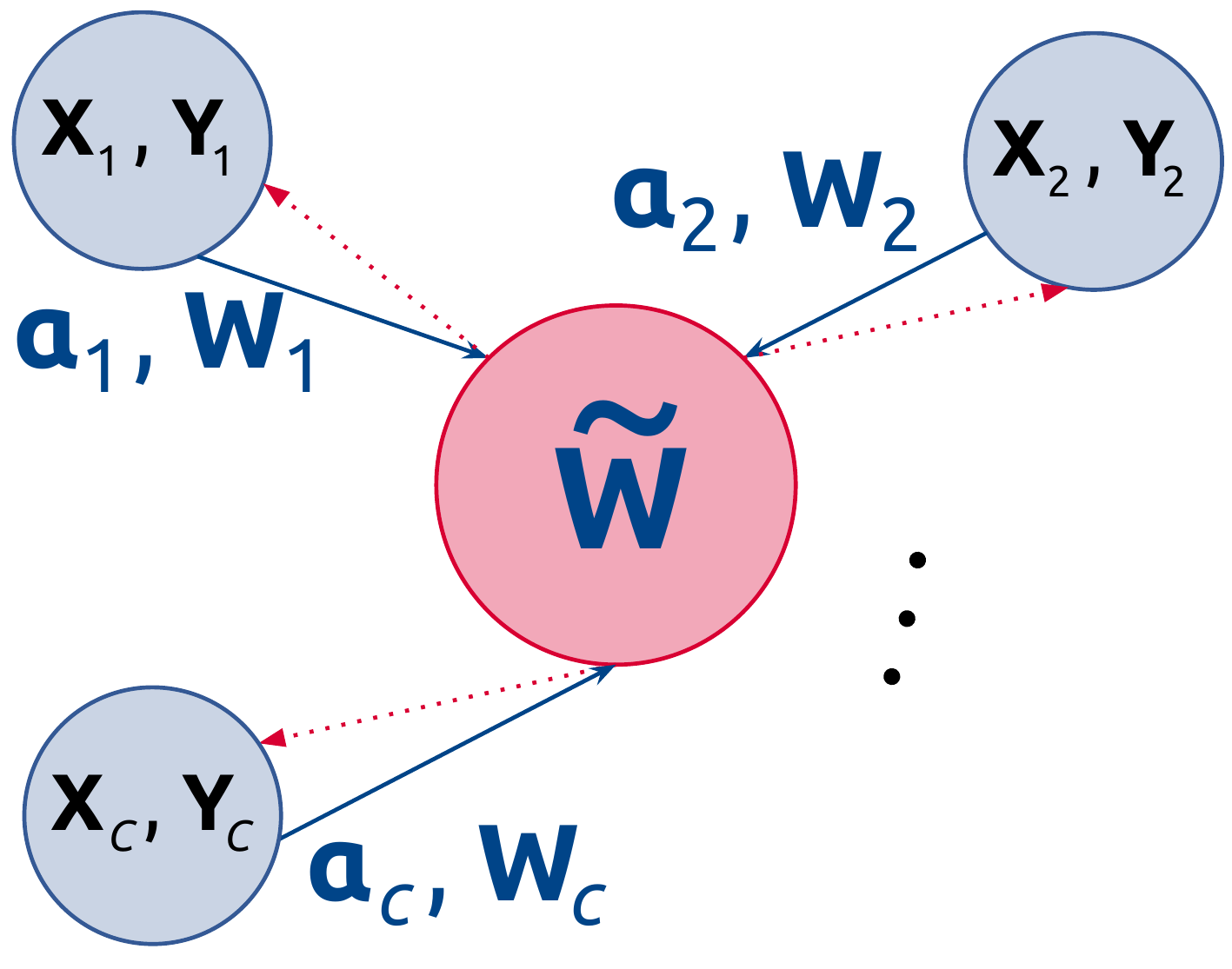}
        \caption{}
    \end{subfigure}
    \begin{subfigure}[b]{0.32\linewidth}
        \includegraphics[width=\linewidth]{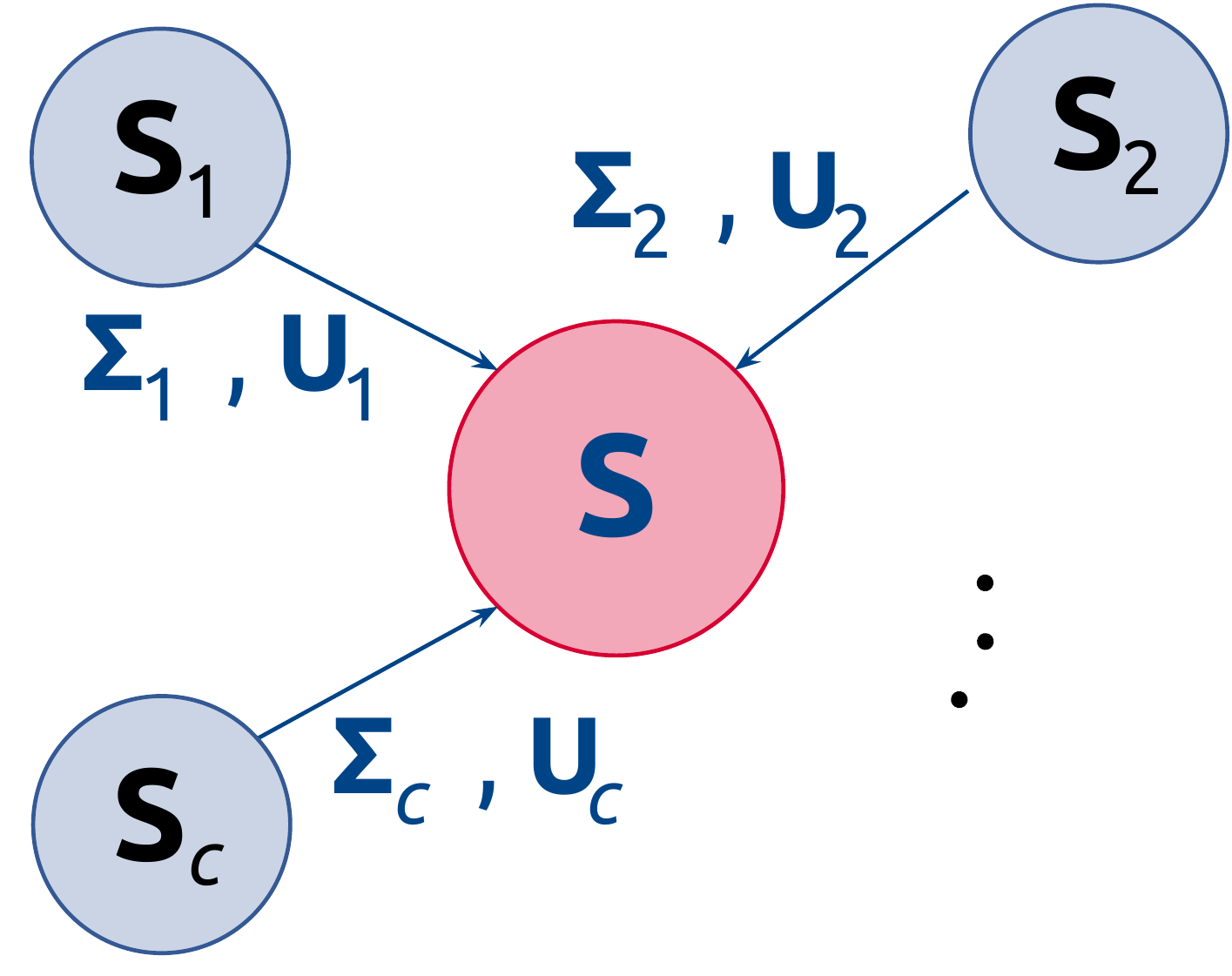}
        \caption{}
    \end{subfigure}
    \caption{Data flow to obtain: (a) the global statistics $\Bar{\mathbf{x}}$ and $\mathbf{\sigma}$, (b) the shared parameter matrix $\widehat{\mathbf{W}}$ to correct from covariates and (c) the approximated global covariance matrix $\mathbf{S}$. Red node: master; blue nodes: local centers. Arrows denote the data flows from centers (blue) and from the master (red).}
    \label{fig:pipeline}
\end{figure}

The proposed framework is illustrated in Figure \ref{fig:pipeline} and discussed in section~\ref{sec:framework}. It is based on three main steps: 1) data standardization, 2) correction from confounding factors and 3) variability analysis.

Data standardization is a data pre-processing step, aiming to enhance the stability of the analysis and easing the comparison across features. In practice, each feature is mapped to the same space by centering data feature-wise to zero-mean and by scaling to unit standard deviation. However, this is ideally performed with respect to the statistics from the whole study (\textit{global statistics}). This issue is addressed by proposing a distributed standardization method in section~\ref{sec:standardization}.

Confounding factors have a \textit{biasing} effect on the data. To correct for this bias, it is usually assumed a linear effect of the confounders  $\widehat{\mathbf{X}} = \mathbf{Y}\mathbf{W}$, that must be estimated and removed. However, for a distributed scenario, computing $\mathbf{W}$ is not straightforward, since the global data matrix cannot be computed. We propose in section~\ref{sec:correction} to use \textit{Alternating Direction Method of Multipliers} (ADMM) to estimate a matrix $\widetilde{\mathbf{W}}$ shared among centers, closely approximating $\mathbf{W}$. In particular, we show that $\widetilde{\mathbf{W}}$ can be estimated in a federated way, without sharing local data $\mathbf{X}_c$ nor covariates $\mathbf{Y}_c$. 

Finally, through federated principal component analysis (fPCA), we obtain a low dimensional representation of the full data without ever sharing any center's individual information $\mathbf{X}_c,\mathbf{Y}_c$ (section~\ref{sec:pca}).

\subsection{Federated Analysis Framework}
\label{sec:framework}

\subsubsection{Standardization}
\label{sec:standardization}
The mean and standard deviation vectors can be initialized to $\Bar{\mathbf{x}}_0 = 0$ and $\Bar{\mathbf{\sigma}}_0 = 0$. They can be iteratively updated with the information of each center by following standard forms  \cite{Welford1962}, by simply transmitting the quantities $\Bar{\mathbf{x}}_c$ and $\sigma_c$ from center to center.
For each center the scaled data is denoted as $\widehat{\mathbf{X}}_c$ and keeps the dimensions of $\mathbf{X}_c$.

\subsubsection{Correction from confounding factors}
\label{sec:correction}
Under the assumption of a linear relationship between data and confounders, the parameters matrix $\mathbf{W}$ can be estimated via \textit{ordinary least squares}, through the minimization of the error function $f(\mathbf{W}) = \norm{\mathbf{Y} - \widehat{\mathbf{X}}\mathbf{W}}^2$.

In a distributed setting, this approach can be performed locally in each center, ultimately leading to $C$ independent solutions. However, this would introduce a bias in the correction, as covariates are accounted for differently across centers.

To solve this issue, we propose to constrain the local solutions to a global one shared across centers. In this way, the subsequent correction can be consistently performed with respect to the estimated global parameters. Thus, we can formulate the problem of constrained regression via ADMM \cite{Boyd2010}.

For a given error function $f_c(\mathbf{W}_c) = \norm{\mathbf{Y}_c - \widehat{\mathbf{X}}_c\mathbf{W}_c}^2$ associated with each center $c$ and constrained to a estimated global matrix of weights $\widetilde{\mathbf{W}}$ we can pose:
\begin{eqnarray}
&&\textrm{minimize} \, \sum_{c=1}^{C} f_c(\mathbf{W}_c), \quad
\textrm{subject to} \, \mathbf{W}_c = \widetilde{\mathbf{W}},\quad \forall c. \nonumber \label{eq:admm_constrain}
\end{eqnarray}

As this is a constrained minimization problem, the extended Lagrangian can be calculated as a combination of the parameters from each center (eqn.~\ref{eq:lagrangian}).
\begin{multline}
    L_\rho(\mathbf{W}, \widetilde{\mathbf{W}}, \mathbf{\alpha}) = \sum_{c=1}^{C} \Big( f_c(\mathbf{W}_c) + \big\langle \mathbf{\alpha}_c, \mathbf{W}_c - \widetilde{\mathbf{W}} \big\rangle \\ + \frac{\rho}{2}\norm{\mathbf{W}_c - \widetilde{\mathbf{W}}}_2^2 \Big)
    \label{eq:lagrangian}
\end{multline}
Where $\rho$ is a penalty factor (or \textit{dual update step length}) regulating the minimization step length for $\mathbf{W}$ and $\widetilde{\mathbf{W}}$. $\mathbf{\alpha}$ is a \textit{dual variable} to decouple the optimization of $\mathbf{W}$ and $\widetilde{\mathbf{W}}$.

Optimization is performed as follows: i) Each center independently calculates the local parameters $\mathbf{W}_c$ and $\mathbf{\alpha}_c$ (eqn.~\ref{eq:minw} and ~\ref{eq:alpha}); ii) the parameters $\mathbf{W}_c$ and $\mathbf{\alpha}_c$ are shared to estimate the global parameters $\widetilde{\mathbf{W}}$ (eqn.~\ref{eq:minw_tilde}). We note that this last step is performed without sharing either local data or covariates. The parameters $\widetilde{\mathbf{W}}$ are subsequently re-transmitted to the centers and the whole procedure is iterated until convergence:
\begin{multline}
    \mathbf{W}_c^{(k+1)} := \arg\min_{\mathbf{W}_c} L_\rho(\mathbf{W}_c, \widetilde{\mathbf{W}}^{(k)}, \mathbf{\alpha}_c^{(k)}) \\
    = \Big(\widehat{\mathbf{X}}_c' \widehat{\mathbf{X}}_c + \frac{\rho}{2}\mathbf{I} \Big)^{-1} \Big( \widehat{\mathbf{X}}_c' \mathbf{Y}_c - \frac{1}{2}\mathbf{\alpha}_c^{(k)} + \frac{\rho}{2}\tilde{\mathbf{W}}_c^{(k)} \Big)
    \label{eq:minw}
\end{multline}
\begin{equation}
\mathbf{\alpha}_{c}^{(k+1)} := \mathbf{\alpha}_c^{(k)} + \rho \big( \mathbf{W}_c^{(k+1)} - \widetilde{\mathbf{W}}^{(k+1)}\big) \label{eq:alpha}
\end{equation}
\begin{multline}
\widetilde{\mathbf{W}}^{(k+1)} := \arg\min_{\widetilde{\mathbf{W}}} L_\rho(\mathbf{W}_c^{(k+1)}, \widetilde{\mathbf{W}}, \mathbf{\alpha}_c^{(k)}) = \frac{1}{C} \sum_c^C \bigg( \frac{\mathbf{\alpha}_c^{(k)}}{\rho} + \mathbf{W}_c^{(k+1)}\bigg)
\label{eq:minw_tilde}    
\end{multline}

After convergence, $\widetilde{\mathbf{W}}$ is shared across centers, and used to consistently account for covariates by subtracting their effect from the structural data to obtain the corrected observation matrix: $
    \mathbf{E}_c = \widehat{\mathbf{X}}_c - \mathbf{Y}_c\widetilde{\mathbf{W}}.
$

\subsubsection{Federated PCA (fPCA)}
\label{sec:pca}
Principal components analysis (PCA) is a standard approach for dimensionality reduction assuming that the largest amount of information is contained in the directions $\mathbf{U}$ (components) of greater variability. Data can be thus represented by projecting on the low-dimensional space spanned by the main components: $\widehat{\mathbf{E}} = \mathbf{E}\mathbf{U}$.

From the eigen-value decomposition of the global covariance matrix $\mathbf{S} = \mathbf{U} \mathbf{\Sigma}^2 \mathbf{U}'$, the first $m$-eigen-modes $\mathbf{U}=(\mathbf{u}_j)_{j=1}^m$ provide a low-dimensional representation of the overall variation in $\mathbf{E}$. In our federated setting, we note that $\mathbf{S}$ is the algebraic sum of the \textit{local covariance matrices} $\mathbf{S} = \mathbf{EE}' = \sum_{c=1}^{S} \mathbf{E}_c\mathbf{E}_c'$. Based on this observation, Lorenzi \textit{et} al. proposed to share only the eigen-modes and values of the covariance matrix of each center avoiding the access to individual data \cite{Lorenzi2017}.
However, sharing the local-covariance-matrices can still be prohibitive as the dimension is $(N_{\textrm{features}} \times N_{\textrm{features}})$. For this reason, it was proposed to further reduce the dimensionality of the problem by sharing only the principal eigen-components associated with the local covariance matrices: $
    \mathbf{S} \approx \sum_{c=1}^{C} \mathbf{U}_c \mathbf{\Sigma}_c^2 \mathbf{U}_c'
    \label{eq:svd_local}
$.
From the practical point of view, computing the eigen-components can be efficiently performed by solving the eigen-problem associated with the matrix $(\mathbf{X}_c \mathbf{X}_c')^2$ which is usually of much smaller dimension $(N_c \times N_c)$ \cite{Worsley2005}.

In what follows, the number of components shared across centers is automatically defined by fixing a threshold of 80\% on the associated \textit{explained variability} contained in $\mathbf{\Sigma}_c$. 

\section{Experiments}
\label{sec:experiments}
\begin{table*}[t]
\resizebox{\textwidth}{!}{%
\begin{tabular}{l|cccc|cc|c|c}
\hline
\textbf{\textbf{Database (total)}} & \multicolumn{4}{c|}{\textbf{ADNI (802)}}                  & \multicolumn{2}{c|}{\textbf{MIRIAD (68)}} & \textbf{PPMI (232)} & \textbf{UK Biobank (208)} \\ \hline
\textbf{Group}                     & HC           & MCInc        & MCIc         & AD           & HC            & AD               & PD                                   & HC               \\
\textbf{N (females)}               & 109 (115)    & 62 (119)     & 78 (130)     & 89 (100)     & 11 (12)       & 26 (19)          & 85 (147)                             & 116 (92)         \\
\textbf{Age $\pm$ sd}              & 75.79 (4.99) & 74.93 (7.72) & 74.54 (7.09) & 75.19 (7.48) & 69 (7.18)     & 69.17 (7.06)     & 60.69 (8.95)                         & 60.72 (7.52)     \\ \hline
\end{tabular}
}
\caption{Data used in this study. Each study here represents an independent center. The centers are jointly analyzed through the federated analysis proposed in Section \ref{sec:framework}.}
\label{tab:dataset}
\end{table*}

\subsection{Synthetic Data}
\label{sec:exp_synthetic}
\begin{figure}
    \centering
    \includegraphics[width=\linewidth]{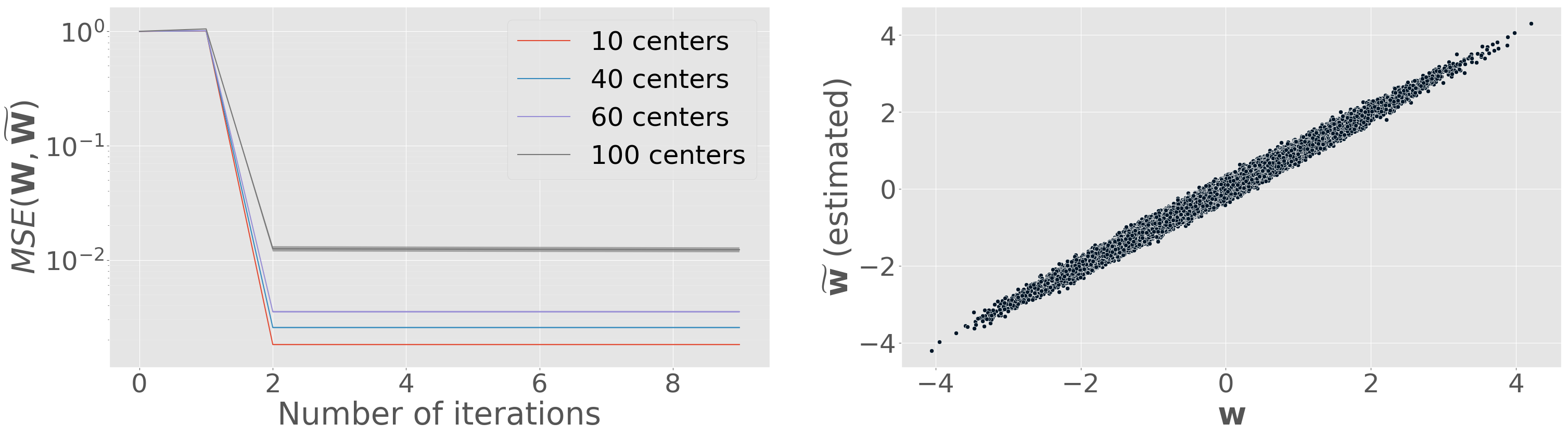}
    \includegraphics[width=\linewidth]{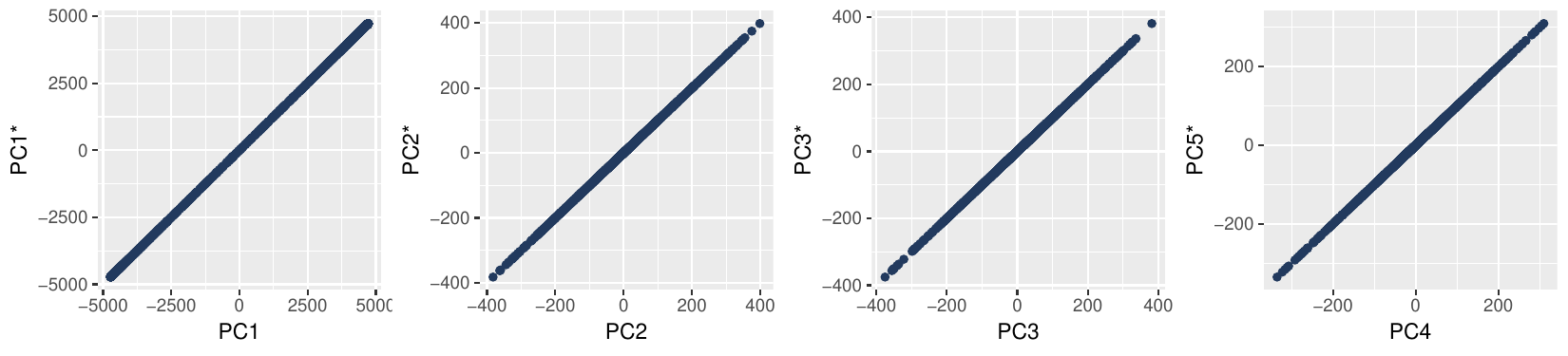}
    \caption{\textbf{Top-left:} Mean square error (MSE) between $\mathbf{W}$ and $\widetilde{\mathbf{W}}$ for different numbers of centers. $N=2400$, $N_{\textrm{features}} = 50,000$ and $\dim(\mathbf{y}) = 20$. \textbf{Top-right:} Single-column of $\mathbf{W}$ vs $\widetilde{\mathbf{W}}$ for $C=100$. \textbf{Bottom:} Principal components (PC) vs federated ones (PC*) for 100 centers.}
    \label{fig:admm_convergence_synt}
\end{figure}

We randomly generated $\mathbf{Y}$ and $\mathbf{W}$ matrices. The data matrix was subsequently computed as $\mathbf{X} = \mathbf{Y}\mathbf{W}$, and corrupted with  Gaussian noise $\mathcal{N}(0,\sigma)$, with $\sigma$ set to 20\% of $\lVert\mathbf{X}\rVert$. Then, $\mathbf{X}$ and $\mathbf{Y}$ were split in $C$ centers of equal sample size. Our federated framework was then applied for each scenario across 200 folds, and convergence analyzed as shown in Figure~\ref{fig:admm_convergence_synt}.
\subsection{Real Data: Neuroimaging}
\label{sec:exp_real}
\textbf{Data.} T1-weighted MRI scans at baseline were analyzed from several research  databases (table~\ref{tab:dataset}). In total, we included data for 455 controls (HC), 181 with non-progressive MCI (MCInc), 208 progressive (MCIc), 234 Alzheimer's disease (AD), 232 with Parkinson's disease (PD).

\textbf{Feature extraction.}
ENIGMA Shape Analysis was applied to the MRI data of each center \cite{Wade2015, Roshchupkin2016}. In our analysis we extracted: a) radial distance (an approximate measure of thickness) and, b) the $\log$ of the Jacobian determinant (surface area dilation/contraction) for each vertex of the following subcortical regions: hippocampi, amygdalae, thalami, pallidum, caudate nuclei, putamen and accumbens nuclei. The overall data dimension is of 54,240 features.

\textbf{Federated analysis.} Each database of table~\ref{tab:dataset} was modeled as an independent center. ${Sex}$, ${Age}$ and ${Age}^2$ were used to correct the vertex-wise shape data according to \ref{sec:standardization} and \ref{sec:correction}. For ADMM, convergence was ensured through 10 iterations. Finally, the analysis of the variability was performed according to \ref{sec:pca}.


\begin{figure}[!htbp]
    \centering
    \includegraphics[height=0.29\linewidth]{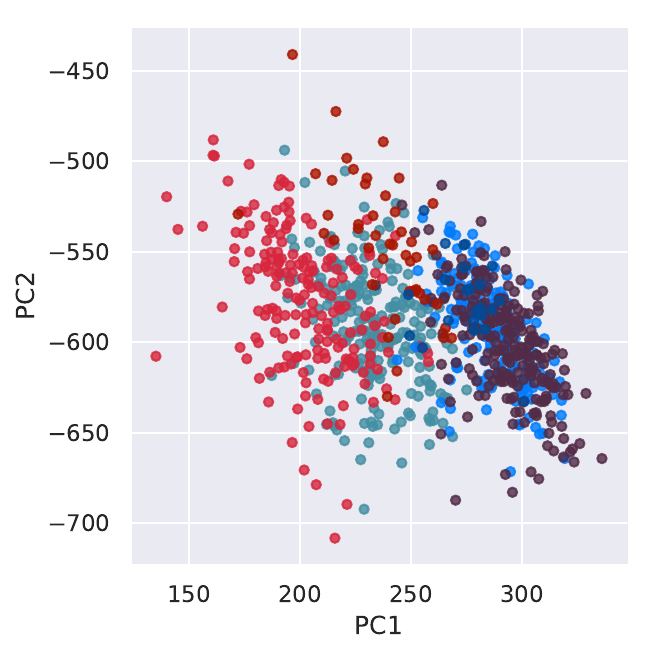}
    \includegraphics[height=0.29\linewidth]{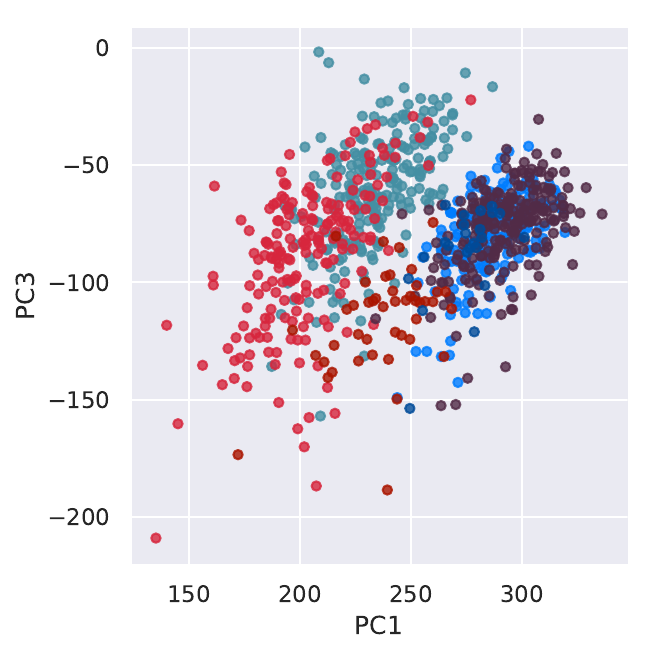}
    \includegraphics[height=0.29\linewidth]{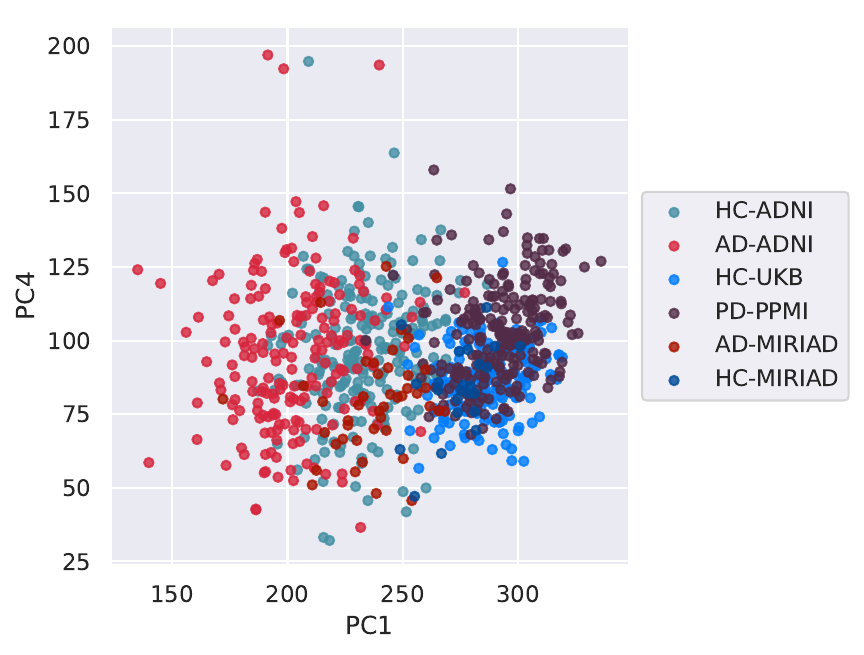}
    
    \includegraphics[height=0.29\linewidth]{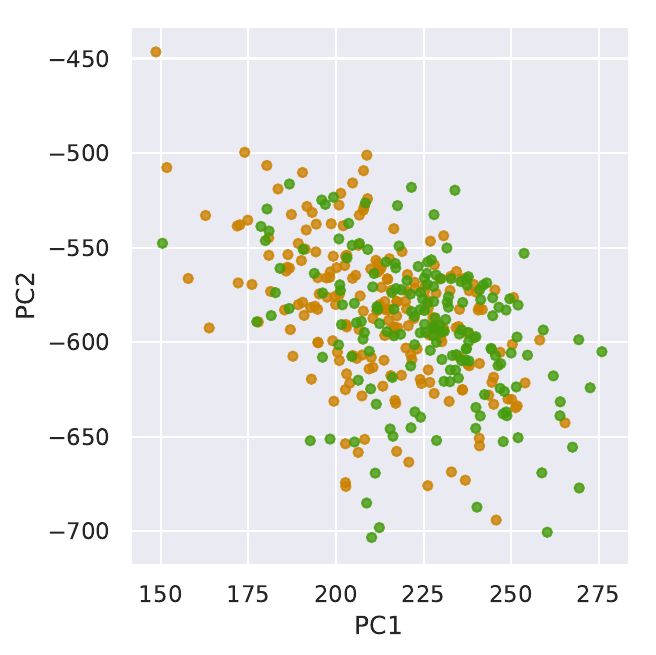}
    \includegraphics[height=0.29\linewidth]{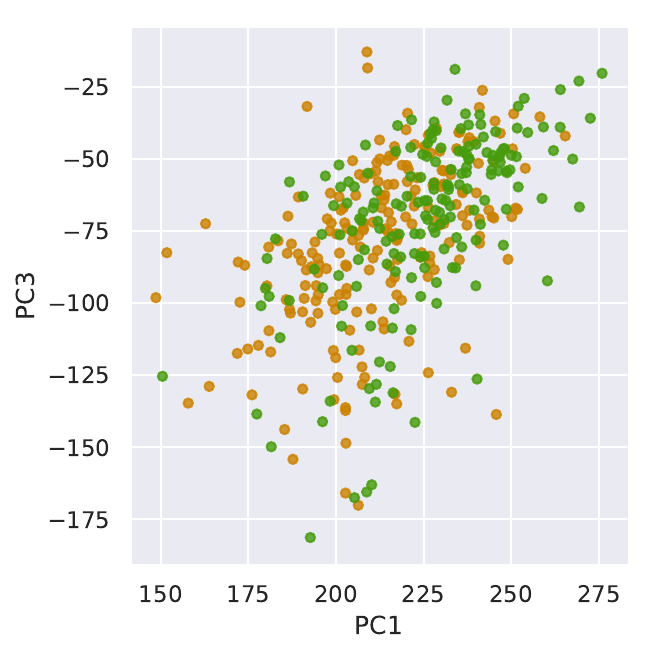}
    \includegraphics[height=0.29\linewidth]{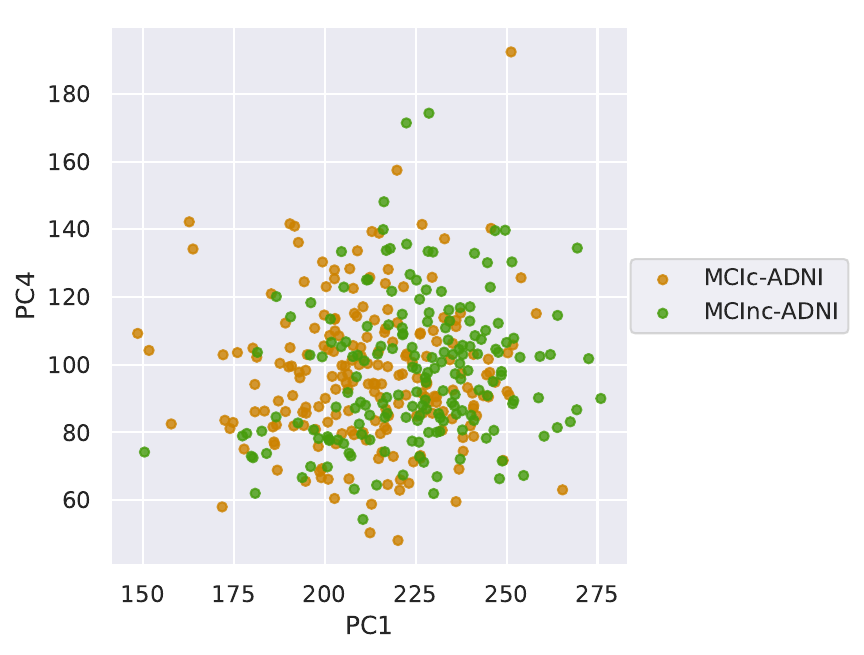}
    \caption{Data projected on the first 4 components. AD vs controls from different centers (top). MCI progressive and stable from ADNI (bottom). Federated PCA was performed on the whole data obtained from the 4 centers (table \ref{tab:dataset}).}
    \label{fig:projection_pca}
\end{figure}

\begin{figure}
    \centering
    \includegraphics[width=0.49\linewidth]{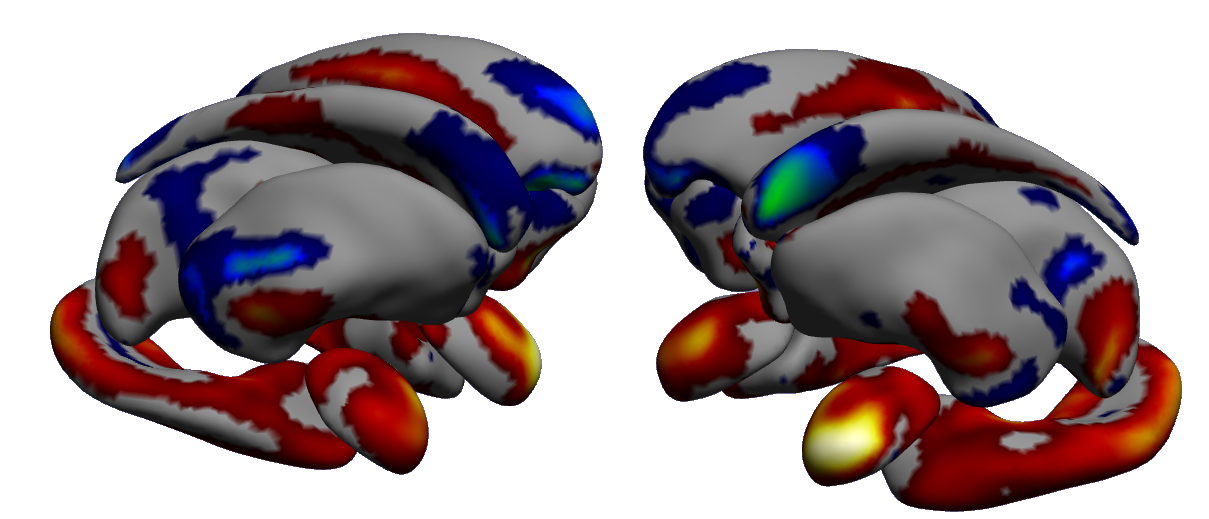}
    \includegraphics[width=0.49\linewidth]{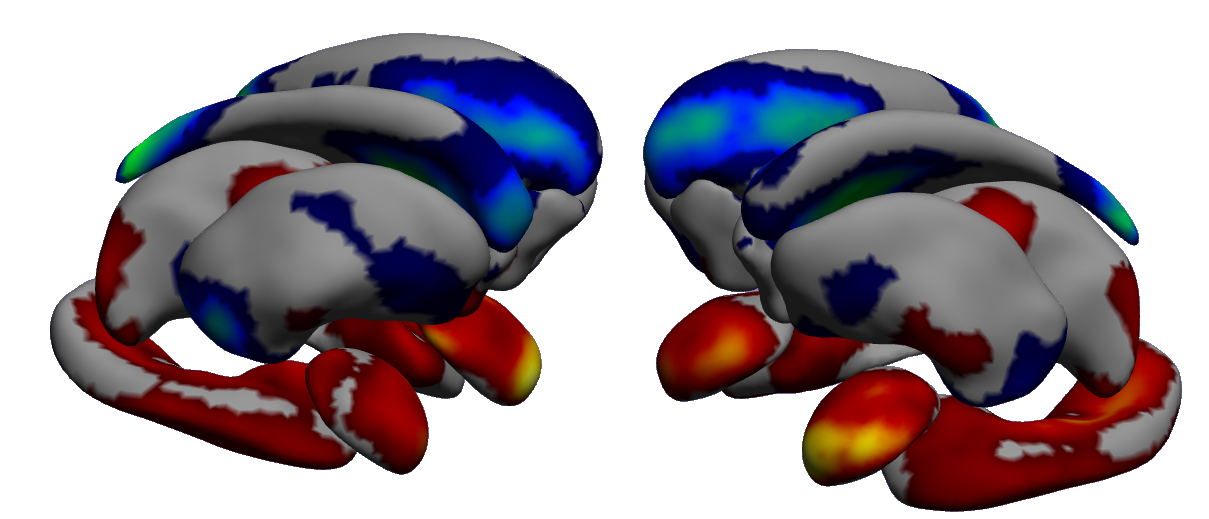}
    \caption{First principal component estimated with the proposed federated framework. The component maps prevalently hippocampi and amigdalae. \textbf{Left:} Thickness. \textbf{Right:} Log-Jacobians.}
    \label{fig:subcortical_mapping}
\end{figure}

\textbf{Results.}
The projection in the latent space spanned by the federated principal components is shown in Figure~\ref{fig:projection_pca}. To ease visualization, the projection for MCI converters and those who remained stable is shown in the bottom panel. Figure ~\ref{fig:subcortical_mapping} shows the weight maps associated to the first principal component. We note that principal components 1 to 3 identify a variability from healthy to AD consistent across centers. Moreover, healthy ADNI participants are in between the AD subjects and the rest of the population. This result may denote some residual effect of $Age$ on the resulting imaging features, even after correction. Interestingly, the issue of ``leaking'' spurious variability of confounders after correction has been already reported in a number of multi-centric studies, and is matter of ongoing research \cite{westfall2016statistically,smith2018statistical}. Finally we note that PD subjects are generally similar to the healthy individuals with respect to the modelled subcortical information. 

\section{Conclusions}
\label{sec:conclusions}
In this work we proposed, tested, and validated a fully consistent framework for federated analysis of distributed biomedical data. Further developments of this study will extend the proposed analysis to large-scale imaging genetics data, such as in the context of the ENIGMA meta-study.

\section{Acknowledgments}
This work was supported by the French government, through the UCAJEDI Investments in the Future project managed by the National Research Agency (ANR) with the reference number ANR-15-IDEX-01 (project Meta- ImaGen). A.A. holds an Medical Research Council eMedLab Medical Bioinformatics Career Development Fellowship. This work was supported by the Medical Research Council [grant number MR/L016311/1]. Additional support is from NIH grants RF1AG051710, R01MH116147 and R56AG058854.\\

This research has been conducted using the UK Biobank Resource under Application Number 20576 (PI Nicholas Ayache).\\

The full list of acknowledgments for the data providers is available at \url{https://hal.inria.fr/hal-01895800/document}.

\bibliographystyle{unsrt}
\bibliography{references.bib}

\begin{thebibliography}{10}

\bibitem{Truffert2017}
Patrick Truffert.
\newblock {Les m{\'{e}}ta-analyses : {\'{e}}l{\'{e}}ments de lecture critique}.
\newblock {\em N{\'{e}}onatologie : bases scientifiques}, pages 15--24, jan
  2017.

\bibitem{Sun2018}
Junfeng Sun et~al.
\newblock {Meta-analysis of Clinical Trials}.
\newblock {\em Principles and Practice of Clinical Research}, pages 317--327,
  jan 2018.

\bibitem{Thompson2014}
Paul~M. Thompson et~al.
\newblock {The ENIGMA Consortium: Large-scale collaborative analyses of
  neuroimaging and genetic data}.
\newblock {\em Brain Imaging and Behavior}, 8(2):153--182, 2014.

\bibitem{baker2015large}
Bradley~T Baker, Rogers~F Silva, Vince~D Calhoun, Anand~D Sarwate, and Sergey~M
  Plis.
\newblock Large scale collaboration with autonomy: Decentralized data ica.
\newblock In {\em 2015 IEEE 25th International Workshop on Machine Learning for
  Signal Processing (MLSP)}, 2015.

\bibitem{ming2017coinstac}
Jing Ming, Eric Verner, Anand Sarwate, Ross Kelly, Cory Reed, Torran Kahleck,
  Rogers Silva, Sandeep Panta, Jessica Turner, Sergey Plis, et~al.
\newblock Coinstac: Decentralizing the future of brain imaging analysis.
\newblock {\em F1000Research}, 6, 2017.

\bibitem{plis2016}
Sergey~M. Plis, Anand~D. Sarwate, Dylan Wood, Christopher Dieringer, Drew
  Landis, Cory Reed, Sandeep~R. Panta, Jessica~A. Turner, Jody~M. Shoemaker,
  Kim~W. Carter, Paul Thompson, Kent Hutchison, and Vince~D. Calhoun.
\newblock Coinstac: A privacy enabled model and prototype for leveraging and
  processing decentralized brain imaging data.
\newblock {\em Frontiers in Neuroscience}, 10:365, 2016.

\bibitem{Lorenzi2017}
Marco Lorenzi et~al.
\newblock {Secure multivariate large-scale multi-centric analysis through
  on-line learning: an imaging genetics case study}.
\newblock page 1016016, 1 2017.

\bibitem{Lorenzi2018}
Marco Lorenzi et~al.
\newblock {Susceptibility of brain atrophy to TRIB3 in Alzheimer’s disease,
  evidence from functional prioritization in imaging genetics}.
\newblock {\em Proceedings of the National Academy of Sciences},
  115(12):3162--3167, 3 2018.

\bibitem{Welford1962}
B.~P. Welford.
\newblock {Note on a Method for Calculating Corrected Sums of Squares and
  Products}.
\newblock {\em Technometrics}, 4(3):419, 8 1962.

\bibitem{Boyd2010}
Stephen Boyd.
\newblock {Distributed Optimization and Statistical Learning via the
  Alternating Direction Method of Multipliers}.
\newblock {\em Foundations and Trends{\textregistered} in Machine Learning},
  3(1):1--122, 2010.

\bibitem{Worsley2005}
Keith~J. Worsley et~al.
\newblock {Comparing functional connectivity via thresholding correlations and
  singular value decomposition}.
\newblock {\em Philosophical Transactions of the Royal Society B: Biological
  Sciences}, 360(1457):913--920, 2005.

\bibitem{Wade2015}
Benjamin~S.C. Wade et~al.
\newblock {Mapping abnormal subcortical brain morphometry in an elderly HIV +
  cohort}.
\newblock {\em NeuroImage: Clinical}, 9:564--573, 2015.

\bibitem{Roshchupkin2016}
Gennady~V. Roshchupkin et~al.
\newblock {Heritability of the shape of subcortical brain structures in the
  general population}.
\newblock {\em Nature Communications}, 7:1--8, 2016.

\bibitem{westfall2016statistically}
Jacob Westfall and Tal Yarkoni.
\newblock Statistically controlling for confounding constructs is harder than
  you think.
\newblock {\em PloS one}, 11(3):e0152719, 2016.

\bibitem{smith2018statistical}
Stephen~M Smith and Thomas~E Nichols.
\newblock {Statistical challenges in “Big Data” human neuroimaging}.
\newblock {\em Neuron}, 97(2):263--268, 2018.

\end{thebibliography}


@article{ming2017coinstac,
  title={COINSTAC: Decentralizing the future of brain imaging analysis},
  author={Ming, Jing and Verner, Eric and Sarwate, Anand and Kelly, Ross and Reed, Cory and Kahleck, Torran and Silva, Rogers and Panta, Sandeep and Turner, Jessica and Plis, Sergey and others},
  journal={F1000Research},
  volume={6},
  year={2017},
  publisher={Faculty of 1000 Ltd}
}

@ARTICLE{plis2016,
    AUTHOR={Plis, Sergey M. and Sarwate, Anand D. and Wood, Dylan and Dieringer, Christopher and Landis, Drew and Reed, Cory and Panta, Sandeep R. and Turner, Jessica A. and Shoemaker, Jody M. and Carter, Kim W. and Thompson, Paul and Hutchison, Kent and Calhoun, Vince D.},   
    TITLE={COINSTAC: A Privacy Enabled Model and Prototype for Leveraging and Processing Decentralized Brain Imaging Data}, 
    JOURNAL={Frontiers in Neuroscience},      
    VOLUME={10},      
    PAGES={365},     
    YEAR={2016},      
    URL={https://www.frontiersin.org/article/10.3389/fnins.2016.00365},       
    DOI={10.3389/fnins.2016.00365},      
    ISSN={1662-453X},   
}

@article{gazula2018decentralized,
  title={Decentralized Analysis of Brain Imaging Data: Voxel-Based Morphometry and Dynamic Functional Network Connectivity},
  author={Gazula, Harshvardhan and Baker, Bradley T and Damaraju, Eswar and Plis, Sergey M and Panta, Sandeep R and Silva, Rogers F and Calhoun, Vince D},
  journal={Frontiers in neuroinformatics},
  volume={12},
  year={2018},
  publisher={Frontiers Media SA}
}

@inproceedings{baker2015large,
  title={Large scale collaboration with autonomy: Decentralized data ICA},
  author={Baker, Bradley T and Silva, Rogers F and Calhoun, Vince D and Sarwate, Anand D and Plis, Sergey M},
  booktitle={2015 IEEE 25th International Workshop on Machine Learning for Signal Processing (MLSP)},
  year={2015}
}



@article{westfall2016statistically,
  title={Statistically controlling for confounding constructs is harder than you think},
  author={Westfall, Jacob and Yarkoni, Tal},
  journal={PloS one},
  volume={11},
  number={3},
  pages={e0152719},
  year={2016},
  publisher={Public Library of Science}
}

@article{smith2018statistical,
  title={{Statistical challenges in “Big Data” human neuroimaging}},
  author={Smith, Stephen M and Nichols, Thomas E},
  journal={Neuron},
  volume={97},
  number={2},
  pages={263--268},
  year={2018},
  publisher={Elsevier}
}

@article{Lorenzi2017,
    title = {{Secure multivariate large-scale multi-centric analysis through on-line learning: an imaging genetics case study}},
    year = {2017},
    booktitle = {Developments in X-Ray Tomography XI},
    author = {Lorenzi, Marco and others},
    month = {1},
    pages = {1016016},
    publisher = {SPIE},
    url = {http://proceedings.spiedigitallibrary.org/proceeding.aspx?doi=10.1117/12.2256799 https://www.spiedigitallibrary.org/conference-proceedings-of-spie/10391/2273320/In-situ-observation-of-polymer-blend-phase-separation-by-x/10.1117/12.2273320.full},
    isbn = {9781510612396},
    doi = {10.1117/12.2256799},
    keywords = {in-situ observation, phase separation, polymer blend, talbot-lau interferometer, x-ray phase tomography}
}

@article{Welford1962,
    title = {{Note on a Method for Calculating Corrected Sums of Squares and Products}},
    year = {1962},
    journal = {Technometrics},
    author = {Welford, B. P.},
    number = {3},
    month = {8},
    pages = {419},
    volume = {4},
    url = {https://www.jstor.org/stable/1266577?origin=crossref},
    doi = {10.2307/1266577},
    issn = {00401706}
}

@article{Lorenzi2018,
    title = {{Susceptibility of brain atrophy to TRIB3 in Alzheimer’s disease, evidence from functional prioritization in imaging genetics}},
    year = {2018},
    journal = {Proceedings of the National Academy of Sciences},
    author = {Lorenzi, Marco and others},
    number = {12},
    month = {3},
    pages = {3162--3167},
    volume = {115},
    url = {http://www.pnas.org/lookup/doi/10.1073/pnas.1706100115},
    doi = {10.1073/pnas.1706100115},
    issn = {0027-8424},
    pmid = {29511103}
}

@article{Thompson2014,
    title = {{The ENIGMA Consortium: Large-scale collaborative analyses of neuroimaging and genetic data}},
    year = {2014},
    journal = {Brain Imaging and Behavior},
    author = {Thompson, Paul M. and others},
    number = {2},
    pages = {153--182},
    volume = {8},
    isbn = {1931-7565 (Electronic){\textbackslash}r1931-7557 (Linking)},
    doi = {10.1007/s11682-013-9269-5},
    issn = {19317565},
    pmid = {24399358},
    keywords = {Consortium, GWAS, Genetics, MRI, Meta-analysis, Multi-site}
}

@article{Boyd2010,
    title = {{Distributed Optimization and Statistical Learning via the Alternating Direction Method of Multipliers}},
    year = {2010},
    journal = {Foundations and Trends{\textregistered} in Machine Learning},
    author = {Boyd, Stephen},
    number = {1},
    pages = {1--122},
    volume = {3},
    url = {http://arxiv.org/abs/1408.2927 http://www.nowpublishers.com/article/Details/MAL-016},
    isbn = {9781627480031},
    doi = {10.1561/2200000016},
    issn = {1935-8237},
    pmid = {17504609},
    arxivId = {1408.2927}
}

@article{Worsley2005,
    title = {{Comparing functional connectivity via thresholding correlations and singular value decomposition}},
    year = {2005},
    journal = {Philosophical Transactions of the Royal Society B: Biological Sciences},
    author = {Worsley, Keith J. and others},
    number = {1457},
    pages = {913--920},
    volume = {360},
    isbn = {0962-8436 (Print)},
    doi = {10.1098/rstb.2005.1637},
    issn = {09628436},
    pmid = {16087436},
    keywords = {Connectivity, Correlation, Singular value decomposition, fMRI}
}

@article{Malone2013,
    title = {{MIRIAD-Public release of a multiple time point Alzheimer's MR imaging dataset}},
    year = {2013},
    journal = {NeuroImage},
    author = {Malone, Ian B. and others},
    month = {4},
    pages = {33--36},
    volume = {70},
    publisher = {Academic Press},
    url = {https://www.sciencedirect.com/science/article/pii/S105381191201230X?via
    isbn = {1095-9572 (Electronic){\textbackslash}r1053-8119 (Linking)},
    doi = {10.1016/j.neuroimage.2012.12.044},
    issn = {10538119},
    pmid = {23274184},
    keywords = {Alzheimer's, Database, Imaging, Longitudinal, MRI}
}



@article{Sun2018,
abstract = {Meta-analysis has become a popular approach for summarizing a large number of clinical trials and resolving discrepancies raised by these trials. In this chapter, we introduce the general procedures for meta-analysis: formulating the question, defining eligibility, identifying studies, abstracting data, statistical analysis, and reporting the results. One key issue determining whether studies can be combined is the extent of heterogeneity among individual studies. We review graphical and statistical tools for assessing heterogeneity, describing the fixed-effect and random-effect models commonly used in meta-analysis, and providing some general recommendations regarding when fixed-effect or random-effect approach is appropriate. Publication bias is an inherent issue with meta-analysis, since studies (especially smaller ones) with “negative” results are frequently unpublished. Funnel plot, Begg and Mazumdar's rank correlation, and Egger regression are useful tools for assessing publication bias. As an illustration of the concepts discussed, we apply meta-analysis techniques to studies examining the use of antiinflammatory therapies in sepsis.},
author = {Sun, Junfeng and others},
doi = {10.1016/B978-0-12-849905-4.00022-8},
isbn = {9780128499054},
journal = {Principles and Practice of Clinical Research},
month = {jan},
pages = {317--327},
publisher = {Academic Press},
title = {{Meta-analysis of Clinical Trials}},
url = {https://www.sciencedirect.com/science/article/pii/B9780128499054000228},
year = {2018}
}

@incollection{Hoffman2015,
abstract = {It is often possible to combine the results of several small trials to obtain a more secure conclusion, but this must be done with care to avoid bias. This chapter discusses how to make these analyses.},
author = {Hoffman, Julien I.E.},
booktitle = {Biostatistics for Medical and Biomedical Practitioners},
doi = {10.1016/B978-0-12-802387-7.00036-6},
isbn = {9780128023877},
month = {jan},
pages = {645--653},
publisher = {Academic Press},
title = {{Meta-analysis}},
url = {https://www.sciencedirect.com/science/article/pii/B9780128023877000366 http://linkinghub.elsevier.com/retrieve/pii/B9780128023877000366},
year = {2015}
}

@article{Truffert2017,
author = {Truffert, Patrick},
doi = {10.1016/B978-2-294-73742-8.00003-0},
isbn = {9782294737428},
journal = {N{\'{e}}onatologie : bases scientifiques},
month = {jan},
pages = {15--24},
publisher = {Elsevier Masson},
title = {{Les m{\'{e}}ta-analyses : {\'{e}}l{\'{e}}ments de lecture critique}},
url = {https://www.sciencedirect.com/science/article/pii/B9782294737428000030},
year = {2017}
}


@article{Wade2017,
abstract = {High-dimensional shape descriptors (HDSD) are useful for modeling subcortical brain surface morphometry. Though HDSD is a useful basis for disease biomarkers, its high dimensionality requires careful treatment in its application to machine learning to mitigate the curse of dimensionality. We explored the use of HDSD feature sets by comparing the performance of two feature selection approaches, Regularized Random Forest (RRF) and LASSO, to no feature selection (NFS). Each feature set was applied to three classifiers: Random Forest (RF), Support Vector Machines (SVM) and Na{\"{i}}ve Bayes (NB). Paired feature-selection-classifier approaches were 10-fold cross-validated on two diagnostic contrasts: Alzheimer's disease and mild cognitive impairment, both relative to controls across varying sample sizes to evaluate their robustness. LASSO aided classification efficiency, however, RRF and NFS afforded more robust performances. Performance varied considerably by classifier with RF being most stable. We advise careful consideration of performance-efficiency tradeoffs in choosing feature selection strategies for HDSD.},
author = {Wade, Benjamin S.C. and others},
doi = {10.1016/j.patcog.2016.09.034},
isbn = {9783319248875},
issn = {00313203},
journal = {Pattern Recognition},
keywords = {Biomarker,Brain,Feature selection,Shape analysis,Subcortical},
month = {mar},
pages = {731--739},
publisher = {Pergamon},
title = {{Machine learning on high dimensional shape data from subcortical brain surfaces: A comparison of feature selection and classification methods}},
url = {https://www.sciencedirect.com/science/article/pii/S0031320316302953},
volume = {63},
year = {2017}
}

@article{Wade2015,
abstract = {Over 50{\%} of HIV + individuals exhibit neurocognitive impairment and subcortical atrophy, but the profile of brain abnormalities associated with HIV is still poorly understood. Using surface-based shape analyses, we mapped the 3D profile of subcortical morphometry in 63 elderly HIV + participants and 31 uninfected controls. The thalamus, caudate, putamen, pallidum, hippocampus, amygdala, brainstem, accumbens, callosum and ventricles were segmented from high-resolution MRIs. To investigate shape-based morphometry, we analyzed the Jacobian determinant (JD) and radial distances (RD) defined on each region's surfaces. We also investigated effects of nadir CD4 + T-cell counts, viral load, time since diagnosis (TSD) and cognition on subcortical morphology. Lastly, we explored whether HIV + participants were distinguishable from unaffected controls in a machine learning context. All shape and volume features were included in a random forest (RF) model. The model was validated with 2-fold cross-validation. Volumes of HIV + participants' bilateral thalamus, left pallidum, left putamen and callosum were significantly reduced while ventricular spaces were enlarged. Significant shape variation was associated with HIV status, TSD and the Wechsler adult intelligence scale. HIV + people had diffuse atrophy, particularly in the caudate, putamen, hippocampus and thalamus. Unexpectedly, extended TSD was associated with increased thickness of the anterior right pallidum. In the classification of HIV + participants vs. controls, our RF model attained an area under the curve of 72{\%}.},
author = {Wade, Benjamin S.C. and others},
doi = {10.1016/j.nicl.2015.10.006},
file = {:user/ssilvari/home/.local/share/data/Mendeley Ltd./Mendeley Desktop/Downloaded/Wade et al. - 2015 - Mapping abnormal subcortical brain morphometry in an elderly HIV cohort.pdf:pdf},
isbn = {9781479923748},
issn = {22131582},
journal = {NeuroImage: Clinical},
keywords = {Classification,HIV,MRI,Random forest,Shape analysis,Subcortical},
pages = {564--573},
pmid = {26413207},
publisher = {The Authors},
title = {{Mapping abnormal subcortical brain morphometry in an elderly HIV + cohort}},
url = {http://dx.doi.org/10.1016/j.nicl.2015.10.006},
volume = {9},
year = {2015}
}

@article{Roshchupkin2016,
abstract = {The volumes of subcortical brain structures are highly heritable, but genetic underpinnings of their shape remain relatively obscure. Here we determine the relative contribution of genetic factors to individual variation in the shape of seven bilateral subcortical structures: the nucleus accumbens, amygdala, caudate, hippocampus, pallidum, putamen and thalamus. In 3,686 unrelated individuals aged between 45 and 98 years, brain magnetic resonance imaging and genotyping was performed. The maximal heritability of shape varies from 32.7 to 53.3{\%} across the subcortical structures. Genetic contributions to shape extend beyond influences on intracranial volume and the gross volume of the respective structure. The regional variance in heritability was related to the reliability of the measurements, but could not be accounted for by technical factors only. These findings could be replicated in an independent sample of 1,040 twins. Differences in genetic contributions within a single region reveal the value of refined brain maps to appreciate the genetic complexity of brain structures.},
author = {Roshchupkin, Gennady V. and others},
doi = {10.1038/ncomms13738},
file = {:user/ssilvari/home/.local/share/data/Mendeley Ltd./Mendeley Desktop/Downloaded/Roshchupkin et al. - 2016 - Heritability of the shape of subcortical brain structures in the general population.pdf:pdf},
issn = {20411723},
journal = {Nature Communications},
pages = {1--8},
pmid = {27976715},
title = {{Heritability of the shape of subcortical brain structures in the general population}},
volume = {7},
year = {2016}
}

@incollection{Jahanshad2018,
abstract = {Large-scale distributed analyses of over 30,000 magnetic resonance imaging scans recently detected common genetic variants associated with the volumes of subcortical brain structures. Scaling up these efforts, still greater computational challenges arise in screening the genome for statistical associations at each voxel in the brain, localizing effects using “image-wide genome-wide” testing (voxelwise genome-wide association studies, vGWASs). Here we benefit from distributed computations at multiple sites to metaanalyze genome-wide image-wide data, allowing private genomic data to stay at the site where it was collected. Site-specific tensor-based morphometry is performed with a custom template for each site, using a multichannel registration. A single vGWAS testing 107 variants against 2million voxels can yield hundreds of terabytes (TB) of summary statistics, which would need to be transferred and pooled for metaanalysis. We propose a two-step method, which reduces data transfer for each site to a subset of single-nucleotide polymorphisms and voxels guaranteed to contain all significant hits.},
author = {Jahanshad, Neda and others},
booktitle = {Imaging Genetics},
doi = {10.1016/B978-0-12-813968-4.00001-8},
isbn = {9780128139684},
month = {jan},
pages = {1--23},
publisher = {Academic Press},
title = {{Chapter One – Multisite Metaanalysis of Image-Wide Genome-Wide Associations With Morphometry}},
url = {https://www.sciencedirect.com/science/article/pii/B9780128139684000018},
year = {2018}
}

\newpage
\section*{Supplementary material: Acknowledgements}
\renewcommand{\thesubsection}{\Alph{subsection}}
\subsection{Funding}
This work was supported by the French government, through the UCAJEDI Investments in the Future project managed by the National Research Agency (ANR) with the reference number ANR-15-IDEX-01 (project Meta- ImaGen). A.A. holds an Medical Research Council eMedLab Medical Bioinformatics Career Development Fellowship. This work was supported by the Medical Research Council [grant number MR/L016311/1]. Additional support is from NIH grants RF1AG051710, R01MH116147 and R56AG058854.

This project has received funding from European Union’s Horizon 2020 research and innovation programme under grant agreement No 847581, the Region Sud and the UCA J.E.D.I.

\subsection{The Alzheimer's Disease Neuroimaging Initiative (ADNI)}
Data used in preparation of this article were obtained from the Alzheimer’s Disease Neuroimaging Initiative (ADNI) database (adni.loni.usc.edu). As such, the investigators within the ADNI contributed to the design and implementation of ADNI and/or provided data but did not participate in analysis or writing of this report. A complete listing of ADNI investigators can be found at: \url{http://adni.loni.usc.edu/wp-content/uploads/how_to_apply/ADNI_ Acknowledgement_List.pdf}

\subsection{The Parkinson's Progression Markers Initiative (PPMI)}
Data used in the preparation of this article were obtained from the Parkinson's Progression Markers Initiative (PPMI) database (www.ppmi-info.org/data). For up-to-date information on the study, visit \url{www.ppmi-info.org}.

PPMI – a public-private partnership – is funded by the Michael J. Fox Foundation for Parkinson’s Research and funding partners, including [list the full names of all of the PPMI funding partners found at \url{www.ppmi-info.org/fundingpartners}.

\subsection{UK Biobank}
This research has been conducted using the UK Biobank Resource under Application Number 20576 (PI Nicholas Ayache). Additional information can be found at: \url{https://www.ukbiobank.ac.uk}.

\end{document}